\documentclass[11pt]{article}

\usepackage[final]{acl}

\usepackage{times}
\usepackage{latexsym}
\usepackage[T1]{fontenc}
\usepackage[utf8]{inputenc}
\usepackage{microtype}
\usepackage{inconsolata}
\usepackage{graphicx}
\graphicspath{{figures/}}
\usepackage{capt-of}
\usepackage{booktabs}
\usepackage{amsmath}
\usepackage{amssymb}
\usepackage{array}
\usepackage{multirow}
\usepackage{xcolor}
\usepackage{tikz}
\usetikzlibrary{arrows.meta,positioning}
\usepackage{url}
\usepackage{subcaption}
\usepackage{placeins}
\usepackage{needspace}
\usepackage{float}
\usepackage{dblfloatfix}

\title{From Relation-Specific Heads to Shared Interventions: Contextual Entrainment in Language Models}

\author{
Jinghan Zheng\thanks{These authors contributed equally to this work.} \\
McGill University \\
\texttt{jinghan.zheng@mail.mcgill.ca}
\And
Jiajun Shi\footnotemark[1] \\
Dalian University of Technology \\
\texttt{violet@mail.dlut.edu.cn}
}
\begin{document}
\maketitle

\begin{abstract}
Large language models rely on prompt context to perform in-context learning, follow formats, and adapt to local evidence. 
This ability, however, also raises a robustness concern: models may reuse  irrelevant tokens from earlier context when answering a later query. 
Prior work identifies this behavior as \emph{contextual entrainment} and finds attention heads whose ablation reduces it, arguing that these heads are largely task- or relation-specific. 
But are the mechanisms behind contextual entrainment truly isolated by relation, or do relation-conditioned searches recover reusable components shared across relations? 
Using factual-distractor relation-completion prompts across 19 lexical-semantic subtypes, we identify attention heads whose ablation reduces entrainment and test how the resulting   head sets transfer recovery ability across exact relations, semantic families, and mixed relation distributions. 
Same-family transfer exceeds different-family transfer, family residuals weaken once globally shared heads are removed, and subtype-specific searches repeatedly rediscover many global heads. 
Clean-completion and context-use probes further suggest that this shared intervention can reduce harmful copying while preserving ordinary relation completion and simple in-context abilities. 
Our findings provide a more precise perspective on contextual entrainment and its mitigation: harmful context following is not best addressed by searching for isolated relation-specific circuits, but by using heterogeneous attack prompts  to reveal reusable intervention targets whose effects are shaped by semantic relation structure.\footnote{Code and experiment notebooks:
\url{https://github.com/billyzheng2410/context-related-hallucination}.}
\end{abstract}

\section{Introduction}

Large language models exhibit a wide range of context-dependent behaviors: they follow instructions, infer task formats from demonstrations, maintain discourse entities, and answer questions based on prompt-local evidence \citep{brown2020fewshot,olsson2022induction}. These abilities, often grouped under in-context learning, entail a sensitivity to previous tokens.

This matters in settings where models must reason over long or heterogeneous inputs, including retrieval-augmented generation, long-context question answering, and tool-augmented interaction \citep{shi2023distracted,yoran2024rag,cuconasu2024noise,wu2024irrelevant,cheng2025stochastic}. 
In such settings, irrelevant context is not always random noise: it can provide structured cues that models incorporate into later predictions. 
\citet{cheng2025stochastic} make this concern concrete through hallucinations in irrelevant-contexts, showing that models can combine abstract class cues with features drawn from the query or context, producing systematic but unreliable answers. 
Our setting studies a token-level version of this problem: the distractor sentence is factually true for another subject, but its answer token becomes misleading evidence for a later, structurally parallel query.

This token-level route is closely related to \emph{contextual entrainment}. 
Following \citet{niu2025llama}, we use this term for cases where tokens that appear earlier in the prompt become more likely to be generated later, even when they do not answer the current query. 
For example:
\begin{quote}
\small
\texttt{The capital of Japan is Tokyo. The capital of China is}
\end{quote}
can increase the probability of the distractor token \texttt{Tokyo}, even though the correct answer is \texttt{Beijing}. The first sentence is true, and the repeated relation frame makes the earlier city name immediately available at the second answer position, where it competes with the correct continuation.

\begin{figure*}[ht]
    \centering
    \includegraphics[width=\textwidth]{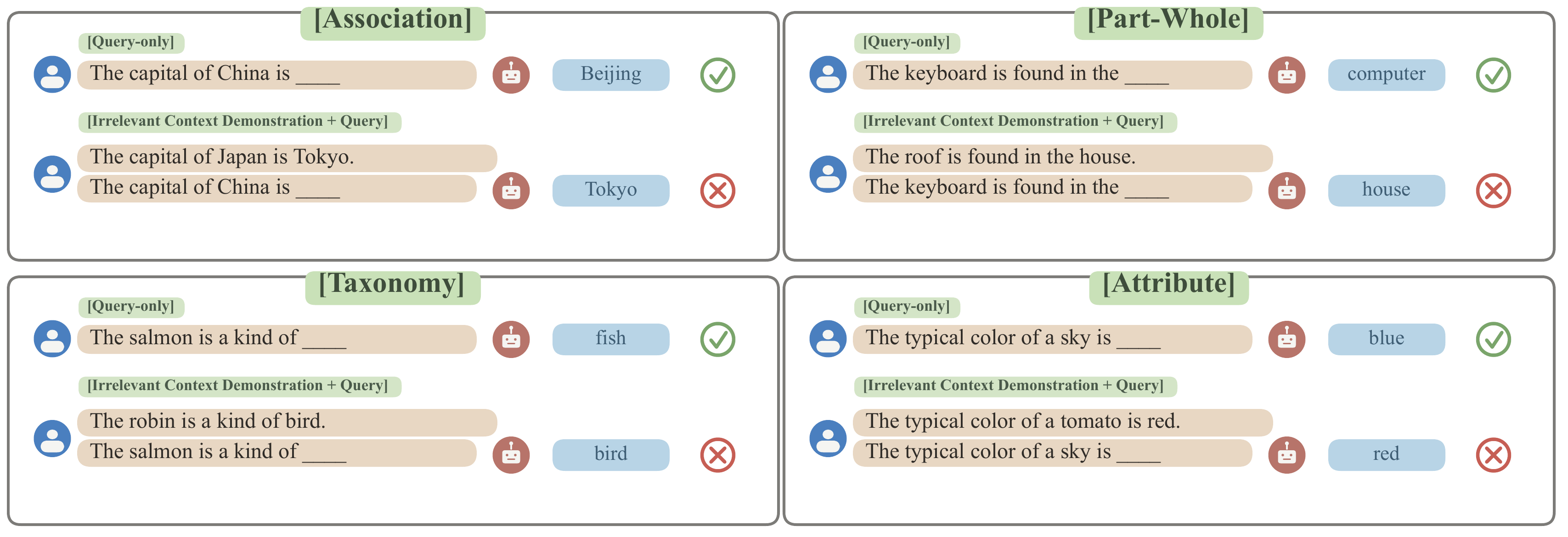}
    \caption{Factual-distractor examples across four lexical-semantic families.}
    \label{fig:context-entrainment}
\end{figure*}

Recent mechanistic work has begun to localize this behavior inside transformer models. 
\citet{niu2025llama} identify attention heads whose ablation reduces contextual entrainment and argue that the recovered head sets are largely task- or relation-specific. They noticed that there existed overlapping attention heads in the result of different relations. However, what this overlap reflects remains unclear. 
Related work on knowledge conflict, however, shows that head-level interventions can sometimes generalize across relations and prompt formats \citep{jin2024cutting}. 
This distinction matters for both interpretation and mitigation. 
If entrainment heads are isolated relation-specific mechanisms, then contextual entrainment is organized into many narrow circuits, and mitigating it may require searching separately for each relation. 
If relation-conditioned searches instead recover overlapping shared components, then apparent task specificity may reflect how a reusable context-following mechanism is selected under different data slices. 
This leaves an open question: when a head set is discovered from a single relation, is it an isolated relation-specific mechanism, or a mixture of components with different scopes?

This question gives a head-level complement to \citet{cheng2025stochastic}: if irrelevant context systematically shapes factual predictions, we ask whether the interventions that reduce this sensitivity are relation-specific or shared across relations. We hypothesize that relation-conditioned searches recover mixtures of shared and relation-sensitive components.
Rather than treating each relation-specific search result as a separate module, we ask whether these head sets contain components that operate at different levels of generality.
We refer to this as a mixed-circuit account of contextual entrainment.
This account makes different predictions from two simpler alternatives.
If entrainment-related heads are purely relation-specific, a head set found on one subtype should transfer little beyond that subtype.
If they are fully general, semantic relatedness should not affect transfer.
The mixed-circuit account predicts the middle case: strongest transfer to the same subtype, weaker but positive transfer within the same semantic family, and weakest transfer across unrelated families.

We test this hypothesis on a controlled lexical-semantic benchmark with 19 relation subtypes grouped into Association, Attribute, Part--Whole, and Taxonomy. 
As illustrated in Figure~\ref{fig:context-entrainment}, each clean relation-completion prompt is paired with an attacked prompt that prepends a true but irrelevant parallel fact from the same subtype. 
The answer token in this prepended fact serves as the distractor: it appears in the immediate context, but it is not the correct continuation for the later query. 
This lets us isolate prompt-local token reuse without introducing an explicit false statement about the queried subject. 
We then search for entrainment-related heads and evaluate how the resulting interventions transfer across same-subtype, same-family, different-family, and mixed-relation conditions.

\paragraph{Contributions.}
We make three contributions.
First, we identify a reusable global intervention component embedded in relation-conditioned head sets, connecting cross-relation transfer with the repeated rediscovery of shared heads.
Second, we provide converging intervention evidence for a mixed-circuit account through comparisons of global, family, and subtype circuits, global-head exclusion, and approximately additive combinations of shared and residual components.
Third, we examine the practical consequences of search scope: in our primary-model experiments, mixed-relation search identifies interventions that reduce factual-distractor entrainment while better preserving clean external relation completion than the evaluated subtype-specific intervention.
\section{Related Work}

\paragraph{Distraction and contextual entrainment.}
LLMs are highly sensitive to prompt context, and prior work shows that irrelevant or misleading evidence can substantially alter model outputs, especially in retrieval-augmented generation and long-context question answering \citep{shi2023distracted,yoran2024rag,cuconasu2024noise,wu2024irrelevant}. \citet{cheng2025stochastic} further show that irrelevant contextual cues can induce structured hallucinations through class-based misgeneralization. \citet{niu2025llama} formalize \emph{contextual entrainment}: tokens that appear earlier in the prompt are more likely to be generated later, even when they are random or semantically irrelevant. They identify attention heads whose ablation reduces entrainment and emphasize relation specificity, while also observing some cross-relation functional overlap.Our study examines how such overlap is organized into reusable shared and relation-sensitive intervention components.

\paragraph{Attention heads as context mechanisms.}
Mechanistic interpretability has shown that attention heads can implement identifiable computations, including induction-like copying, name moving, and task-specific circuits \citep{olsson2022induction,wang2023ioi}. Induction heads are especially relevant here because they implement a form of pattern completion: continuations from earlier contexts are reused at later matching positions \citep{olsson2022induction}. Related work on knowledge conflict further suggests that attention heads can mediate competition between parametric memory and external context, and that head-level interventions can generalize across relations and prompt formats \citep{jin2024cutting}. Together, these findings make it plausible that factual-distractor prompts share a reusable context-following component even when their surface relations differ.

\paragraph{Circuit discovery by differentiable masking.}
A growing line of work uses activation patching, pruning, and differentiable masks to identify causally relevant components in language models. \citet{yu-etal-2025-sheaf} use Gumbel-Sigmoid masking for gradient-based joint computation-graph pruning in language-model circuit discovery, building on continuous relaxations of discrete variables and sparse stochastic gates \citep{maddison2017concrete,jang2017categorical,louizos2018learning}. \citet{niu2025llama} apply a related Gumbel-Sigmoid head-search framework to contextual entrainment. These methods make it possible to move beyond descriptive transfer patterns and ask whether head sets discovered from individual relations reflect isolated mechanisms or repeatedly recovered shared structure.

\paragraph{Shared and distributed mechanisms.}
Broader mechanistic work argues that model mechanisms can be distributed, overlapping, and partly redundant rather than cleanly modular \citep{elhage2022superposition,mcgrath2023hydra}. This matters for head search: low overlap between recovered head sets is not, by itself, evidence for isolated task-specific mechanisms. At the same time, work on relation representations suggests that factual relations are systematically organized in model states \citep{hernandez2024linearity}. Together, these perspectives motivate our question: whether relation-conditioned causal searches recover narrow local effects, or broader shared structure with relation-sensitive modulation.

\section{Methodology}

We first define the attacked relation-completion task and intervention, then
describe how head sets are searched and evaluated, before introducing the
benchmark and transfer conditions used to test our hypothesis.
All main experiments use LLaMA-3.1-8B \citep{dubey2024llama}. 
Interventions are performed by zero-ablating selected attention head outputs during the forward pass, while leaving all model parameters unchanged.

\subsection{Task Definition}

We define contextual entrainment as an attacked relation-completion task with
four entities and answers: a \emph{query subject} \(s_q\), its \emph{gold answer}
\(y_{\mathrm{gold}}\), a different \emph{distractor subject} \(s_d\), and its
\emph{distractor answer} \(y_{\mathrm{dist}}\). Using the country--capital
example in Figure~\ref{fig:context-entrainment}, \texttt{China} is the query
subject, \texttt{Beijing} is the gold answer, \texttt{Japan} is the distractor
subject, and \texttt{Tokyo} is the distractor answer. We use this terminology
throughout the paper.

Each example is built from a relation template \(T(\cdot)\). The clean prompt
queries the answer for the query subject alone, while the attacked prompt
prepends a structurally parallel factual statement about the distractor
subject:
\[
x_{\mathrm{clean}} = T(s_q), \ \
x_{\mathrm{attack}} = T(s_d) + y_{\mathrm{dist}} + T(s_q).
\]
The attacked prompt places a concrete distractor answer in the local context of
the later query, where it competes with the gold answer at the shared answer
position.

We use \emph{factual distractors} rather than counterfactual context. The
prepended statement is true for another subject but irrelevant to the current
query. This design isolates prompt-local token reuse from explicit
memory--context conflict, making the attack weaker but cleaner. Our goal is to
test whether a true earlier answer for one subject interferes with the correct
continuation for another.

\subsection{Gumbel-Sigmoid Head Search}

To identify candidate anti-entrainment circuits, we run a differentiable
head-search procedure in the spirit of \citet{niu2025llama}.Each attention head receives a learnable binary mask $m_i$,
where $m_i=1$ zeros its output-projected contribution to the
residual stream at all sequence positions. We optimize these masks with a Gumbel-Sigmoid straight-through
relaxation \citep{maddison2017concrete,jang2017categorical,louizos2018learning},
which allows gradient-based search over discrete head subsets. We then use the
learned mask scores to define ranked top-$K$ interventions for later analyses.

For attacked prompts, the objective maximizes recovery of the gold-over-distractor margin while encouraging sparsity:
\begingroup
\small
\[
\mathcal{L}
=
-\mathbb{E}_{x\in\mathcal{B}}
\left[
M_{\theta}^{\mathrm{abl}}(x)-M_{\theta}(x)
\right]
+
\frac{\lambda}{|\mathcal{H}|}
\sum_i \sigma(\alpha_i).
\]
\endgroup

After search, the learned mask scores define a head ranking, from which we
export top-\(K\) interventions. We use \(K=80\), selected by a held-out sweep
balancing attack recovery and clean-prompt harm. Additional optimization
details are in Appendix~\ref{app:gumbel_details}.

Subtype-specific, family-shared, and global-shared sets contain
the top-$K$ heads searched on one subtype, a balanced mixture
of subtypes within one family, and a balanced mixture of all
eligible subtypes, respectively.
Family-residual sets exclude global-shared heads from the full
family ranking and retain the top-$K$ remaining heads without
retraining.

\subsection{Evaluation Metrics}

We evaluate interventions by how much they restore the model's preference for
the gold answer over the distractor answer. For a prompt \(x\), let \(\ell(y\mid x)\)
be the model logit for token \(y\) at the answer position. We measure this
preference with the margin
\[
M(x) = \ell(y_{\mathrm{gold}} \mid x) - \ell(y_{\mathrm{dist}} \mid x).
\]
This logit margin directly measures the model's relative preference between the gold and distractor answers, with the softmax normalizer canceling out, and is therefore less noisy than probability differences under ablations that can shift mass to unrelated tokens.

The factual distractor is harmful when it lowers this margin, so we define attack drop as
\[
D = M(x_{\mathrm{clean}}) - M(x_{\mathrm{attack}}).
\]
Because context following is generally useful, we evaluate recovery only on
\emph{valid attacked cases}: the clean prompt must favor the gold answer and the
distractor must reduce the margin by at least \(D \geq 0.5\). This isolates
cases where prompt-local context induces a measurable factual-distractor
effect.

For a head set \(S\), the raw causal effect of zero ablation is
\[
\Delta_S = M_S(x_{\mathrm{attack}}) - M(x_{\mathrm{attack}}),
\]
where \(M_S\) is the margin after ablating \(S\). Positive \(\Delta_S\) means
the intervention shifts the model away from the distractor and toward the gold
answer. Since target subtypes differ in attack strength, we also report
normalized recovery,
\[
R_S = \frac{\Delta_S}{D},
\]
where \(R_S=1\) indicates full recovery of the attack-induced margin loss (e.g., if ablating \(S\) improves the attacked margin by \(\Delta=0.5\) on both \emph{country--capital} (France--Paris) with \(D=0.5\) and \emph{part--whole} (wheel--car) with \(D=2\), then \(R\) is \(1\) vs. \(0.25\)).
Values above \(1\) indicate over-recovery: the intervention increases the
attacked margin by more than the margin loss induced by the distractor. We
therefore interpret such values as intervention effects rather than bounded
percentages. Similar over-recovery effects are reported by
\citet{niu2025llama}.

For transfer analyses, we further remove target-specific recoverability by centering normalized recovery within each target:
\[
\widetilde{R}_{s,t}
=
R_{s,t}
-
\frac{1}{|\mathcal{S}|}
\sum_{s' \in \mathcal{S}} R_{s',t}.
\]
This asks whether a source circuit performs better than the average source
circuit on the same target (e.g., on target \emph{country--capital} (France--Paris), if most sources cluster around \(R\approx0.6\) but a same-family source reaches \(R=0.8\), then \(\widetilde{R}\approx+0.2\) captures above-average transfer rather than target ease). We use \(\widetilde{R}\) for same-family versus
different-family comparisons, while also reporting raw margin changes \(\Delta\).

\subsection{Lexical-Semantic Benchmark}

Having defined the task, intervention, and evaluation metrics in general terms, we now instantiate them on a controlled lexical-semantic benchmark with 19 relation subtypes grouped into four families: Association, Attribute,
Part--Whole, and Taxonomy. Taxonomic, part--whole, and attribute relations are
standard organizing relations in lexical semantics and WordNet-style lexical
databases \citep{cruse1986lexical,miller1995wordnet,fellbaum1998wordnet};
Association covers non-taxonomic factual and functional mappings common in
relation-completion benchmarks \citep{petroni2019lama,hernandez2024linearity}.
Representative examples and statistics are shown in Figure~\ref{fig:context-entrainment} and Table~\ref{tab:benchmark_stats}, respectively. More details are reported in Appendix A.

\subsection{Transfer and Circuit Conditions}

A circuit searched on source subtype \(s\) is evaluated on target subtype \(t\). We group source--target pairs as \emph{same-subtype}, \emph{same-family}, or \emph{different-family} transfer. To analyze recovered circuits, we compare \emph{subtype-specific}, \emph{family-shared}, \emph{global-shared}, and \emph{family-residual} head sets, where family residuals exclude the global-shared top-\(K\) heads from the corresponding family ranking. We also measure head overlap and define frequency-global heads by rediscovery frequency across subtype-specific searches.

\begin{table}[!b]
\centering
\small
\setlength{\tabcolsep}{4pt}
\begin{tabular*}{\columnwidth}{@{\extracolsep{\fill}}lrrrrr}
\toprule
\textbf{Family} & \textbf{Sub.} & \textbf{Pairs} & \textbf{Valid test} & \textbf{Rate} & $\mathbf{\bar{D}}$ \\
\midrule
Association & 4 & 163 & 441/480 & 91.9 & 3.92 \\
Attribute & 5 & 244 & 293/600 & 48.8 & 2.69 \\
Part--Whole & 5 & 224 & 492/600 & 82.0 & 4.77 \\
Taxonomy & 5 & 227 & 496/600 & 82.7 & 3.60 \\
\midrule
\textbf{Total} & \textbf{19} & \textbf{858} & \textbf{1722/2280} & \textbf{75.5} & \textbf{3.86} \\
\bottomrule
\end{tabular*}
\caption{
Benchmark statistics by semantic family after tokenizer and valid-attack filtering.
\textbf{Valid test} uses the main filter \(D\geq0.5\); \(\bar{D}\) is the mean attack drop.
}
\label{tab:benchmark_stats}
\end{table}

\section{Results}
The results test whether relation-conditioned head searches identify isolated relation-specific circuits or mixed circuits with shared components.
At first glance, searches on different lexical-semantic subtypes return different top-\(K\) heads, and these differences matter for transfer.
The key result, however, is that this surface-level task dependence masks a shared structure: relation-conditioned searches repeatedly include a global component, with family- and subtype-level variation layered around it.

We organize this pattern using the mixed-circuit hypothesis. For a subtype of relation $r$, a set of heads recovered by relation-specific search can be functionally viewed as
\[
C_r \approx G + F_{\mathrm{fam}(r)} + S_r + \epsilon,
\]
where $G$ is a reusable global context-following component, $F_{\mathrm{fam}(r)}$ is semantic-family modulation, $S_r$ is subtype-specific structure, and $\epsilon$ captures template, optimizer, or factual-completion idiosyncrasies. This decomposition is not intended to be a strict linear factorization of head effects; rather, it provides a guide to interpret transfer, residual, and overlap patterns.

We first verify that the recovered head sets have genuine causal effects on factual-distractor entrainment and that transfer is graded by semantic relatedness (Figure~\ref{fig:semantic_transfer}). We then use transfer, residualization, head overlap, additive intervention tests, and clean-capability probes to distinguish isolated task-specific circuits from the mixed-circuit account.

\subsection{Transfer Is Graded by Semantic Relatedness}
\label{sec:semantic_transfer}
\begin{figure}[t]
    \centering
    \includegraphics[width=\columnwidth]{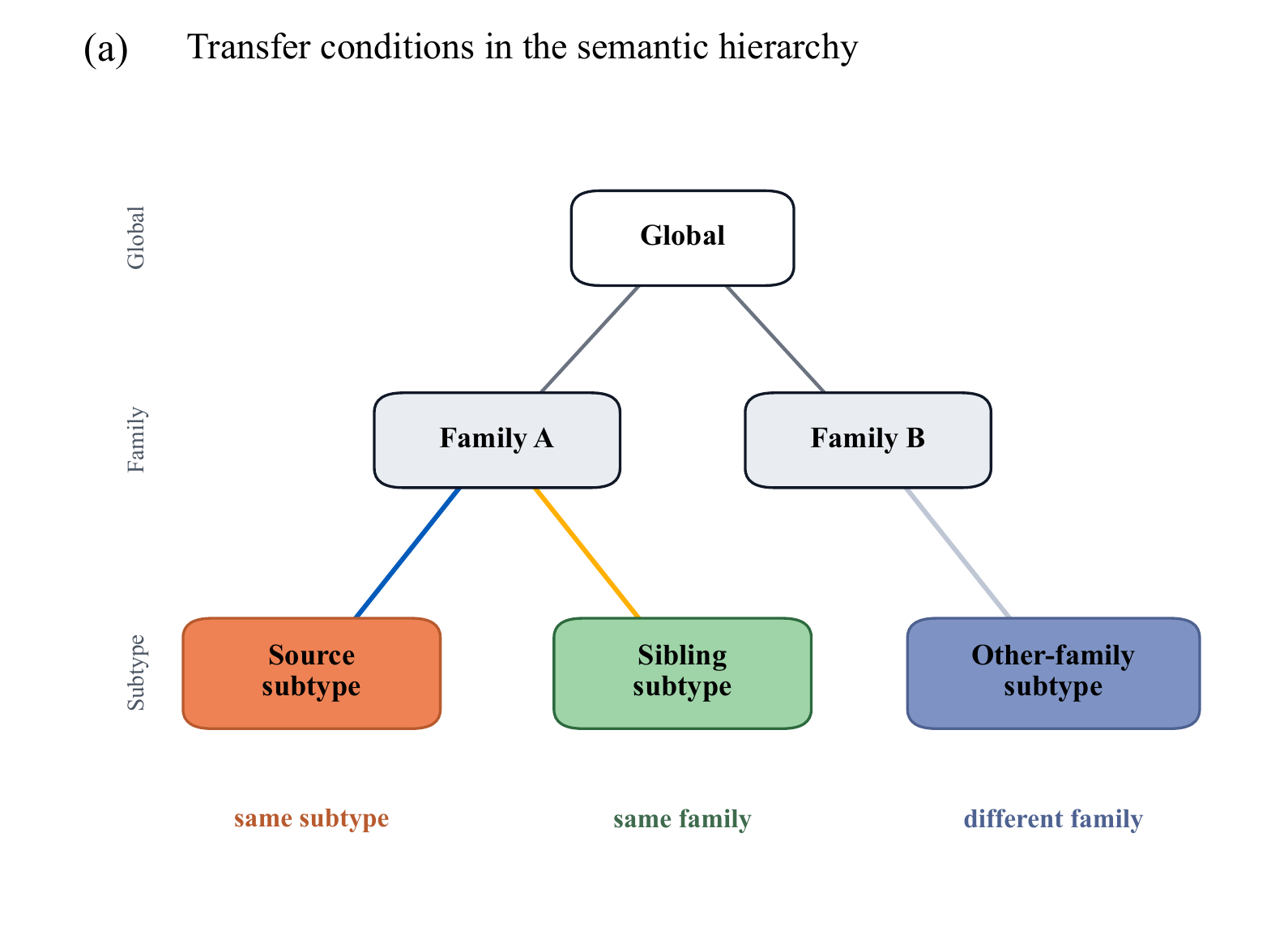}

    \vspace{0.25em}

    \includegraphics[width=\columnwidth]{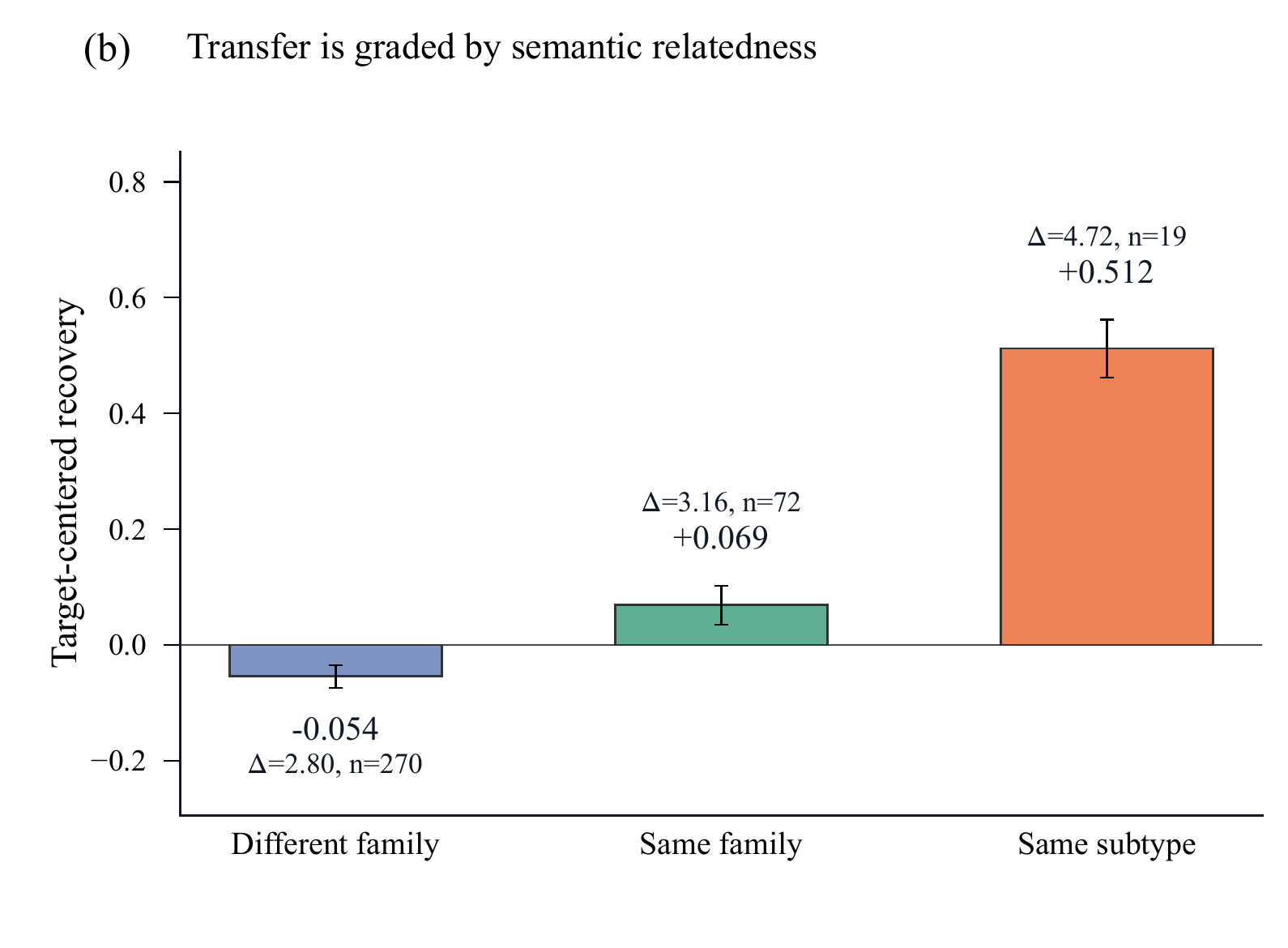}

    \caption{
    Transfer is graded by semantic relatedness. The top panel illustrates the transfer conditions from a source circuit to target relations. The bottom panel reports target-centered recovery, which controls for target-specific recoverability.
    }
    \label{fig:semantic_transfer}
\end{figure}
Figure~\ref{fig:semantic_transfer} summarizes this structure: the transfer matrix falls between two extremes---isolated relation-specific circuits and fully universal heads.

Figure~\ref{fig:semantic_transfer} shows a graded pattern: different-family transfer is weakest, same-family transfer is stronger, and same-subtype transfer is strongest. After target-centering, different-family transfer is below the target average ($-0.054$), same-family transfer is above it ($+0.069$), and same-subtype transfer is much larger ($+0.512$). This hierarchy remains significant under clustered bootstrap, cell-level tests, and family-label permutation controls; full statistics are reported in Appendix~\ref{app:transfer_stats}.

Relation-specific searches therefore do not recover isolated subtype modules, but neither do they recover a fully relation-agnostic circuit. Semantic family and exact subtype both modulate transfer.

\subsection{Family Effects Are Built on Global Shared Heads}
\label{sec:global_family_decomposition}

The graded transfer pattern suggests that semantic relatedness modulates transfer, but it does not by itself determine whether this structure reflects family-specific modules or variations around a shared mechanism. We therefore compare four circuit constructions: circuits searched on a balanced mixture of all active subtypes (\emph{global-shared}), circuits searched only within the target family (\emph{family-shared}), circuits searched on the exact target subtype (\emph{subtype-specific}), and family rankings after removing global-shared top-\(K\) heads (\emph{family residual}).
\begin{table}[t]
\centering
\small
\setlength{\tabcolsep}{4pt}
\begin{tabular*}{\columnwidth}{@{\extracolsep{\fill}}lccccc}
\toprule
\textbf{Circuit} & \textbf{Assoc.} & \textbf{Attr.} & \textbf{P--W} & \textbf{Tax.} & \textbf{Avg.} \\
\midrule
Global shared & $1.502$ & $1.801$ & $1.028$ & $1.531$ & $1.464$ \\
Family shared & $1.486$ & $1.904$ & $0.925$ & $0.943$ & $1.306$ \\
Family residual & $0.881$ & $1.243$ & $0.699$ & $0.997$ & $0.959$ \\
Subtype-specific & $1.606$ & $1.619$ & $1.069$ & $1.361$ & $1.404$ \\
\bottomrule
\end{tabular*}
\caption{
Shared-circuit decomposition. Values are normalized recovery. Family residuals exclude global-shared top-\(K\) heads from the family ranking.
}
\label{tab:shared_decomposition}
\end{table}

Table~\ref{tab:shared_decomposition} supports the mixed-circuit account. The global-shared circuit is the strongest aggregate intervention, with average recovery $\mathbf{1.464}$, while family-shared circuits remain effective but weaker on average ($\mathbf{1.306}$). Subtype-specific circuits are also strong ($\mathbf{1.404}$), indicating that exact relation structure still contributes. Thus, family-conditioned search is useful, but it is not the dominant source of recovery.

The residual comparison provides more direct evidence. Once globally shared heads are removed from the family ranking, average family recovery drops from $\mathbf{1.306}$ to $\mathbf{0.959}$. The residual remains positive, so family structure is not absent. However, the drop shows that much of the apparent family-shared effect is carried by heads that also belong to the global circuit. In the decomposition, $G$ carries the largest recoverable component, while $F_{\mathrm{fam}(r)}$ and $S_r$ account for the remaining semantic variation in transfer.
Thus, same-family transfer is real, but much of this family structure is organized around a reusable global core rather than independent family modules.

\begin{table}[t]
\centering
\small
\setlength{\tabcolsep}{4pt}
\begin{tabular}{lccc}
\toprule
\textbf{Head set} & \textbf{$K=20$} & \textbf{$K=40$} & \textbf{$K=80$} \\
\midrule
Global search & $0.207$ & $0.724$ & $1.464$ \\
Frequency-defined global & $0.364$ & $0.639$ & $1.350$ \\
Frequency $\cap$ global & $0.361$ & $0.788$ & $1.431$ \\
Subtype-specific & $0.301$ & $0.654$ & $1.404$ \\
Family-shared & $0.204$ & $0.551$ & $1.306$ \\
Rare heads & $0.004$ & $0.014$ & $0.082$ \\
\bottomrule
\end{tabular}
\caption{
Frequency-defined global heads recover most of the global circuit's effect, while rarely rediscovered heads do not. Values are normalized recovery.
}
\label{tab:frequency_global}
\end{table}

\subsection{Relation-Specific Searches Rediscover the Global Core}
\label{sec:global_rediscovery}

A strong global circuit could still be an artifact of mixed-relation search. If so, it would not need to appear inside relation-specific circuits. The mixed-circuit account predicts the opposite: the same global heads should be repeatedly rediscovered when searching on individual subtypes.

This prediction is borne out in the overlap statistics. Global-shared circuits overlap with family-shared circuits by 40.25 heads on average, compared with 6.25 expected by random overlap. They overlap with subtype-specific circuits by 32.84 heads on average, and 75 of the 80 global-shared heads appear in at least one subtype-specific top-$K$ circuit. The global circuit is therefore not separate from relation-specific circuits; it is largely made of heads that narrower searches also select.

Rediscovery frequency is itself predictive of causal effect. If we rank heads only by how often they appear in subtype-specific top-$K$ circuits, the resulting frequency-defined global heads recover much of the global-search circuit. At $K=80$, they reach recovery $1.350$, close to the global-search recovery $1.464$, while rarely rediscovered heads recover only $0.082$ (Table~\ref{tab:frequency_global}). This provides head-level evidence that $G$ is already embedded in the subtype circuits $C_r$.

\subsection{The Mixed-Circuit Decomposition Predicts Intervention Effects}
The mixed-circuit account treats a relation-conditioned head set as a combination of a global component, a family-level residual, and a subtype-level residual. 

For each target subtype, we construct disjoint head sets corresponding to the global
component $G$, the family-residual component $F$, and the subtype-residual
component $S$, and evaluate interventions that include different subsets of
these components. We then fit a regression over raw margin recovery:
\[
\Delta = \beta_0 + \beta_G I_G + \beta_F I_F + \beta_S I_S + \epsilon,
\]
where $I_G$, $I_F$, and $I_S$ indicate whether the corresponding component is
ablated.

The fitted coefficients support the mixed-circuit decomposition. The global
component has the largest effect ($\beta_G=2.936$), while family and subtype
residuals also contribute positively ($\beta_F=1.009$, $\beta_S=1.159$). The
additive model explains substantial variance in raw recovery ($R^2=0.572$);
adding all interaction terms barely improves fit ($R^2=0.578$), and additive
and full-interaction predictions are nearly identical ($r=0.995$). Thus, the
components are not independent modules, but their causal effects are well
approximated by an additive decomposition.

A fixed-budget diagnostic supports the same conclusion. An 80-head mixture combining $G_{40}$, $F_{20}$, and $S_{20}$ achieves the strongest recovery among matched 80-head interventions, outperforming pure global, family-only, and subtype-specific alternatives. This indicates that $G$ carries the dominant recoverable effect, while $F_{\mathrm{fam}(r)}$ and $S_r$ add smaller but useful residual contributions. Details in Appendix \ref{app:additive_decomposition}.

\subsection{Global Circuits Preserve Clean Relation Completion}
\label{sec:clean_capability}

A final test asks whether the global circuit reduces entrainment by broadly disrupting relation-completion behavior rather than by selectively counteracting the distractor. If ablating it only damaged general relation completion, then its recovery effect would be less meaningful. We therefore run a clean capability evaluation on external LRE relations, where prompts contain no distractor context.

\begin{figure}[t]
\centering
\includegraphics[width=\columnwidth]{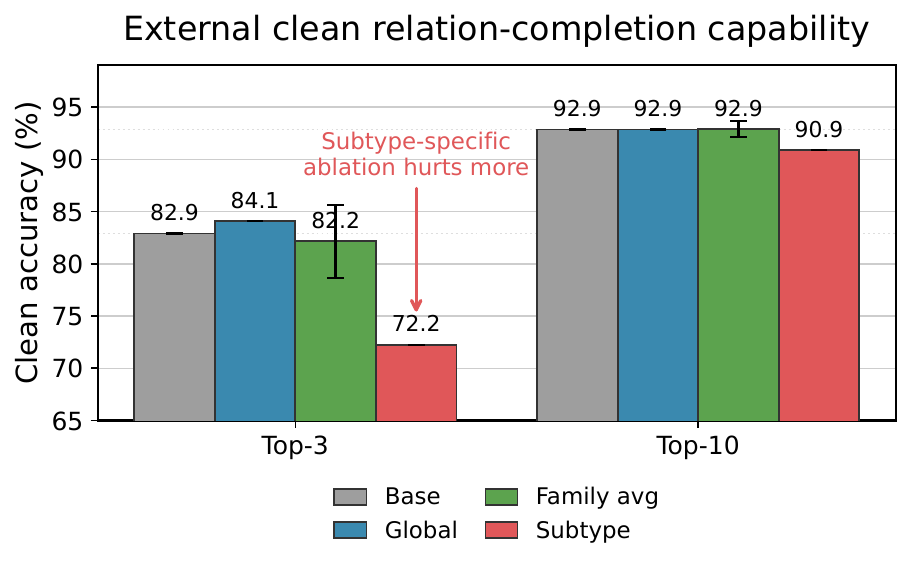}
\caption{
External LRE clean relation-completion capability after ablation.
}
\label{fig:clean_capability}
\end{figure}

Figure~\ref{fig:clean_capability} summarizes the clean-capability probe. The global circuit preserves external relation completion closely, and family-shared circuits remain similarly close to the base model on average. By contrast, subtype-specific circuits cause larger collateral degradation across the 19 subtype-specific interventions. The figure uses the country--capital subtype as an example of a narrow subtype-specific intervention; the full subtype-specific breakdown is reported in Appendix~\ref{app:lre}.

This pattern is consistent with the idea that mixed-relation search is not merely broader, but cleaner. One plausible explanation is that single-relation search can entangle distractor suppression with ordinary factual-completion machinery for that relation. Heterogeneous search instead favors heads whose ablation consistently reduces factual-distractor entrainment across relation environments while better preserving clean relation completion.

\section{Robustness and Reproducibility}

The core empirical pattern is stable under several additional checks: transfer remains semantically graded, and the shared global component remains useful across alternative analyses.

\textbf{Controls:} Uniformly random and same-layer random head controls are near zero or negative, indicating that recovery is not a generic consequence of ablating many heads or ablating heads from sensitive layers (Appendix~\ref{app:transfer_stats}).
\textbf{Thresholds, templates, and seeds:} Same-family transfer remains stronger than different-family transfer across valid-attack thresholds from \(D \geq 0.25\) to \(D \geq 1.0\). Cross-template recovery remains positive, and an eight-subtype three-seed matrix preserves the same qualitative hierarchy.
\textbf{Model replication:} We replicate the main experiments on Mistral-7B \citep{jiang2023mistral} and Qwen2.5-7B \citep{yang2024qwen25} under the same evaluation protocol.
\textbf{Additional probes:} A context-use probe suggests that global ablation does not notably damage ordinary in-context learning ability, as reflected in simple few-shot spelling correction, translation, and in-context mapping tasks. Appendix E further shows that the global effect is not specific to zero ablation via clean-to-attacked activation replacement.

Detailed results are reported in Appendices B--E.

\section{Discussion}

\paragraph{From task-specific heads to mixed circuits.}
Our results refine the task-specific interpretation of contextual entrainment heads proposed by \citet{niu2025llama}. Consistent with prior work, searches on different relation tasks do not return identical head sets. This variability is not surprising: transformer computations are often distributed and redundant, with multiple components able to support overlapping functions or compensate for one another under intervention \citep{elhage2022superposition,mcgrath2023hydra}. The key question is therefore not whether different tasks select exactly the same heads, but whether the selected heads are functionally isolated. Our results suggest that they are not. Relation-conditioned searches repeatedly contain a global component that transfers across semantic families, while family and subtype structure modulate this shared component. 
\begin{figure}[t]
\centering
\includegraphics[width=\columnwidth]{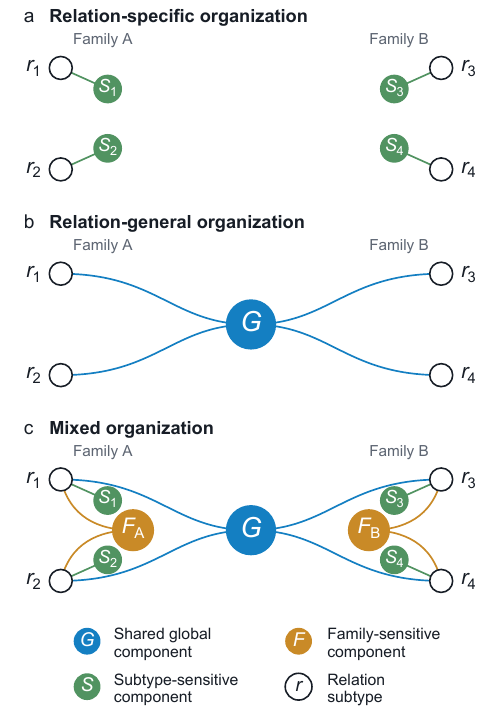}
\caption{Comparison among single-relation, general-relation and mixed-relation search.}

\label{fig:sel_pressure_mini}
\end{figure}

\paragraph{Why a global core is significant.}
Contextual entrainment is behaviorally general: models often reuse prompt-local structure across tasks and relations \citep{olsson2022induction,niu2025llama}. However, this behavioral generality does not by itself imply a shared circuit-level intervention. The same vulnerability could have been implemented by largely separate relation-specific head sets, in which case a circuit found for one relation would transfer little beyond that relation. Our results show instead that relation-conditioned searches repeatedly recover a strong relation-general component. Semantic family and subtype structure still shape the effect, but mainly by modulating this shared component rather than replacing it with separate modular circuits. Thus, the significance of the global core is mechanistic: apparent relation specificity partly reflects different uses of a reusable context-following substrate.

\paragraph{Why mixed search can reduce side effects.}

Single-relation search entangles two components: heads that reduce distractor copying and heads involved in normal relation completion. This explains why subtype-specific circuits are strong but have larger side effects. Heterogeneous search changes the selection pressure: a head is favored only if its ablation consistently reduces distractor entrainment across different relations. The resulting global circuit better preserves ordinary completion while targeting harmful context following.

\paragraph{Implication for RAG and factual context use.}
The global component also suggests a practical direction for settings with heterogeneous factual context. In retrieval-augmented generation, multi-document question answering, dialogue history, and tool-augmented agent traces, the model may see true facts that are useful locally but irrelevant to the current query. A mitigation method therefore cannot simply suppress context use, since factual grounding is the purpose of these systems. Prior relation-specific interventions may be difficult to deploy in such settings, where a single context can mix many relation types and factual roles. In our controlled relation-completion setting, mixed-relation
search identifies reusable interventions that reduce
factual-distractor copying while largely preserving clean
completion.
These findings motivate future evaluation of shared interventions
in RAG and long-context tasks.

\section{Conclusion}

By analyzing entrainment across lexical-semantic relations, we show that context following in LLMs is neither fully relation-specific nor fully relation-general. Head-level intervention and transfer analyses reveal mixed circuits: a reusable global component supports broad context following, while semantic-family and subtype structure shape how this component is used in particular relations. These findings challenge the view that entrainment-related heads form isolated relation-specific modules. Apparent task specificity instead emerges from overlapping shared and relation-sensitive components. For mitigation, this suggests starting from heterogeneous attack data to identify broad reusable interventions, while using relation-specific searches for fine-grained specialization.

\section{Limitations}

Several limitations remain. LLaMA-3.1-8B serves as our primary case study, and corresponding cross-model results for Mistral-7B and Qwen2.5-7B are reported in Appendix~\ref{app:robustness}. Our benchmark is also controlled: prompts use relation-completion templates, and answers are restricted to single tokens so that gold-versus-distractor margins are directly comparable. This design improves interpretability, but it leaves open whether the same mixed-circuit structure appears in more natural long-context or retrieval settings. We did not conduct our experiments using larger LMs due to resource limitations.

Our conclusions are also conditional on valid attacked cases. This filtering is appropriate for studying harmful entrainment, since recovery is meaningful only when the factual distractor actually weakens the gold-over-distractor margin. However, it does not characterize all uses of context following, including cases where context helps the model answer correctly. Finally, head ablation identifies causally useful components, but it does not fully specify the internal algorithm by which the model resolves competition between the gold answer and the distractor. Future work should combine head-level interventions with activation patching or representation-level analyses, and should evaluate stability over more seeds and broader prompt distributions.

\clearpage
\bibliography{references}
\clearpage
\appendix

\section{Benchmark Details}
\label{app:dataset}

Table~\ref{tab:appendix_subtypes} lists the 19 relation subtypes used in the benchmark.
Each subtype contains factual $(x,y)$ pairs and two relation-specific templates.
We filter examples so that both the gold answer and the distractor answer are single tokens under the model tokenizer.

Tables~\ref{tab:appendix_subtypes} and~\ref{tab:appendix_templates} provide the subtype and template details.

\paragraph{Definition of Association.}
We use \textsc{Association} as an operational cover family for relation
subtypes that are not taxonomic, meronymic, or attributive, but are linked by
conventional thematic, functional, institutional, or usage-based knowledge
(e.g., country--capital, profession--workplace, sport--equipment, and
task--tool). This follows computational lexical-semantics work emphasizing
that many useful word-pair relations are broader, non-classical relational
links rather than WordNet-style taxonomic or part--whole relations
\citep{morris-hirst-2004-non,turney-2006-similarity,jurgens-etal-2012-semeval}.
We therefore treat \textsc{Association} as a heterogeneous contrast family,
not as a single classical semantic primitive.
\paragraph{Overlapping family boundaries.}
The family labels are operational groupings rather than mutually exclusive lexical-semantic primitives. Some relation types naturally sit near the boundary between families. Material relations are a clear example: they can be viewed as substance meronymy when the material is treated as a constituent of an object, but also as attribute-value relations when the material is treated as a property of the object. We therefore retain material-based subtypes in both the Part--Whole and Attribute families instead of manually removing these boundary cases. This choice makes the family-transfer test more conservative: overlap between families should blur, rather than sharpen, same-family versus different-family distinctions. The fact that we still observe graded same-subtype, same-family, and different-family transfer under this operational grouping supports the robustness of the semantic-structure result.
\FloatBarrier

\begin{table*}[!tbp]
\centering
\small
\setlength{\tabcolsep}{5pt}
\begin{tabular}{llll}
\toprule
\textbf{Family} & \textbf{Subtype} & \textbf{Example pair} & \textbf{Representative template} \\
\midrule
Association & country--capital & France $\rightarrow$ Paris & The capital of France is Paris. \\
Association & profession--workplace & doctor $\rightarrow$ hospital & The workplace of a doctor is a hospital. \\
Association & sport--equipment & tennis $\rightarrow$ racket & The equipment used in tennis is a racket. \\
Association & task--tool & writing $\rightarrow$ pen & The tool used for writing is a pen. \\
\midrule
Attribute & color & banana $\rightarrow$ yellow & The typical color of a banana is yellow. \\
Attribute & shape & coin $\rightarrow$ round & The typical shape of a coin is round. \\
Attribute & material & window $\rightarrow$ glass & The typical material of a window is glass. \\
Attribute & temperature & ice $\rightarrow$ cold & The typical temperature of ice is cold. \\
Attribute & size & elephant $\rightarrow$ large & The typical size of an elephant is large. \\
\midrule
Part--Whole & component--object & wheel $\rightarrow$ car & The wheel is a component of the car. \\
Part--Whole & member--collection & player $\rightarrow$ team & The player is a member of the team. \\
Part--Whole & portion--whole & slice $\rightarrow$ cake & The slice is a portion of the cake. \\
Part--Whole & material--object & table $\rightarrow$ wood & The table is made of wood. \\
Part--Whole & body-part--body & finger $\rightarrow$ hand & The finger is part of the hand. \\
\midrule
Taxonomy & animal & sparrow $\rightarrow$ bird & The sparrow is a kind of bird. \\
Taxonomy & plant & rose $\rightarrow$ flower & The rose is classified as a flower. \\
Taxonomy & artifact & hammer $\rightarrow$ tool & The hammer is a type of tool. \\
Taxonomy & food & apple $\rightarrow$ fruit & The apple is a type of fruit. \\
Taxonomy & vehicle & sedan $\rightarrow$ car & The sedan is a kind of car. \\
\bottomrule
\end{tabular}
\caption{Relation subtypes and representative templates in the lexical-semantic benchmark.}
\label{tab:appendix_subtypes}
\end{table*}

\begin{table*}[!tbp]
\centering
\small
\setlength{\tabcolsep}{5pt}
\begin{tabular}{lll}
\toprule
\textbf{Subtype} & \textbf{Template 1} & \textbf{Template 2} \\
\midrule
country--capital & The capital of \{x\} is \{y\}. & The capital city of \{x\} is \{y\}. \\
profession--workplace & The workplace of a \{x\} is a \{y\}. & A \{x\} typically works in a \{y\}. \\
sport--equipment & The equipment used in \{x\} is a \{y\}. & People use a \{y\} in \{x\}. \\
task--tool & The tool used for \{x\} is a \{y\}. & People use a \{y\} for \{x\}. \\
\midrule
color & The typical color of a \{x\} is \{y\}. & The color commonly associated with a \{x\} is \{y\}. \\
shape & The typical shape of a \{x\} is \{y\}. & The shape commonly associated with a \{x\} is \{y\}. \\
material & The typical material of a \{x\} is \{y\}. & A \{x\} is typically made of \{y\}. \\
temperature & The typical temperature of \{x\} is \{y\}. & A \{x\} is typically \{y\} in temperature. \\
size & The typical size of a \{x\} is \{y\}. & A \{x\} is typically \{y\} in size. \\
\midrule
component--object & The \{x\} is a component of the \{y\}. & The \{y\} contains a component called a \{x\}. \\
member--collection & The \{x\} is a member of the \{y\}. & The \{y\} includes the member \{x\}. \\
portion--whole & The \{x\} is a portion of the \{y\}. & The \{y\} contains a portion called a \{x\}. \\
material--object & The \{x\} is made of \{y\}. & The material of the \{x\} is \{y\}. \\
body-part--body & The \{x\} is part of the \{y\}. & The \{y\} contains the body part \{x\}. \\
\midrule
animal & The \{x\} is a kind of \{y\}. & The \{x\} is classified as a \{y\}. \\
plant & The \{x\} is a kind of \{y\}. & The \{x\} is classified as a \{y\}. \\
artifact & The \{x\} is a type of \{y\}. & The \{x\} is classified as a \{y\}. \\
food & The \{x\} is a type of \{y\}. & The \{x\} is classified as a \{y\}. \\
vehicle & The \{x\} is a kind of \{y\}. & The \{x\} is classified as a \{y\}. \\
\bottomrule
\end{tabular}
\caption{Prompt templates used for each relation subtype.}
\label{tab:appendix_templates}
\end{table*}

\FloatBarrier
\clearpage
\section{Additional Transfer Statistics}
\label{app:transfer_stats}

\paragraph{Valid-attack filtering.}
Our main experiments define valid attacked cases as examples where the clean prompt favors the gold answer and the factual distractor reduces the gold-over-distractor margin by at least \(D \geq 0.5\).
This retains 1,722 of 2,280 LLaMA-3.1-8B test attacks across 19 relation subtypes.
Table~\ref{tab:threshold_sensitivity} shows that the same-family advantage is stable across nearby threshold choices.

\begin{table}[H]
\centering
\scriptsize
\setlength{\tabcolsep}{3pt}
\resizebox{\columnwidth}{!}{%
\begin{tabular}{lrrrrrr}
\toprule
\textbf{Threshold} &
\textbf{Valid tests} &
\textbf{Same-family} &
\textbf{Diff.-family} &
\textbf{Raw adv.} &
\textbf{Centered adv.} &
\textbf{\(p\)} \\
\midrule
\(D \geq 0.25\) & 1776 & 0.997 & 0.871 & 0.125 & 0.128 & 0.0022 \\
\(D \geq 0.50\) & 1722 & 0.956 & 0.838 & 0.118 & 0.123 & 0.0021 \\
\(D \geq 0.75\) & 1660 & 0.926 & 0.812 & 0.113 & 0.118 & 0.0021 \\
\(D \geq 1.00\) & 1590 & 0.885 & 0.779 & 0.106 & 0.112 & 0.0024 \\
\bottomrule
\end{tabular}%
}
\caption{Valid-attack threshold sensitivity. Same-family transfer remains stronger than different-family transfer across all tested attack-drop thresholds. Same-family and different-family columns report non-target-centered normalized recovery. The centered advantage is computed after target-centering; \(p\) is the cell-level Welch test for the centered same-family versus different-family comparison.}
\label{tab:threshold_sensitivity}
\end{table}

\paragraph{Dependency-aware inference.}
For the same-family advantage test, we use target-centered recovery and additionally run a two-way clustered bootstrap over source and target subtypes.
For each bootstrap sample, we resample source and target subtypes with replacement, reconstruct the corresponding transfer cells, recompute target-centered recovery, and then recompute the same-family minus different-family advantage.

\begin{table}[H]
\centering
\small
\setlength{\tabcolsep}{6pt}
\resizebox{\columnwidth}{!}{%
\begin{tabular}{lrrrr}
\toprule
\textbf{Statistic} & \textbf{Observed} & \textbf{Median} & \textbf{95\% CI} & \textbf{\(p\)} \\
\midrule
Same-family advantage & 0.123 & 0.115 & [0.003, 0.236] & 0.024 \\
\bottomrule
\end{tabular}%
}
\caption{Two-way clustered bootstrap over source and target subtypes. The statistic is the target-centered same-family minus different-family advantage.}
\label{tab:cluster_bootstrap}
\end{table}

\paragraph{Random baselines.}
Table~\ref{tab:random_baselines_full} reports the full random and same-layer random controls.
Each cell reports non-target-centered normalized recovery, with raw attacked-margin change in parentheses.

\begin{table}[H]
\centering
\small
\setlength{\tabcolsep}{6pt}
\resizebox{\columnwidth}{!}{%
\begin{tabular}{lccc}
\toprule
\textbf{Method} & \textbf{Diff.-family} & \textbf{Same-family} & \textbf{Same-subtype} \\
\midrule
Gumbel & 0.693 (2.711) & 1.081 (3.737) & 1.407 (5.418) \\
Random & -0.167 (-0.695) & -0.089 (-0.395) & -0.123 (-0.458) \\
Same-layer random & -0.202 (-0.659) & -0.117 (-0.349) & -0.213 (-0.818) \\
\bottomrule
\end{tabular}%
}
\caption{Random and same-layer random controls.}
\label{tab:random_baselines_full}
\end{table}
\paragraph{Graded transfer detail.}
Table~\ref{tab:target_centered_transfer} reports target-centered recovery and raw margin gains for the three transfer conditions.
\begin{table}[H]
\centering
\small
\setlength{\tabcolsep}{5pt}
\resizebox{\columnwidth}{!}{%
\begin{tabular}{lrr}
\toprule
\textbf{Transfer condition} & \textbf{Target-centered recovery} & \textbf{Raw \(\Delta\)} \\
\midrule
Different-family & -0.054 & 2.80 \\
Same-family & +0.069 & 3.16 \\
Same-subtype & +0.512 & 4.72 \\
\bottomrule
\end{tabular}%
}
\caption{Target-centered transfer by semantic relatedness. Target-centering removes target-specific recoverability before comparing source circuits.}
\label{tab:target_centered_transfer}
\end{table}

\clearpage
\FloatBarrier
\section{Robustness and Cross-Model Replication}
\label{app:robustness}
This appendix reports robustness checks for template variation, seed stability, and cross-model replication. 
\paragraph{Template robustness.}
Cross-template recovery remains positive but weaker than same-template recovery:
Association drops from \(1.671\) to \(1.488\), Attribute from \(1.855\) to \(1.610\), Part--Whole from \(0.801\) to \(0.690\), and Taxonomy from \(1.721\) to \(0.968\).

\paragraph{Seed stability.}
In an eight-subtype mini transfer matrix over three seeds, same-family transfer remains positive \((0.071, 0.276, 0.259)\), same-subtype transfer is larger \((0.510, 0.726, 0.631)\), and different-family transfer remains negative \((-0.097, -0.167, -0.148)\).

\paragraph{Cross-model replication.}
Table~\ref{tab:cross_model_transfer} reports non-target-centered transfer recovery for all three models under valid-attack filtering.
Using the same protocol, we completed the five core analyses on Mistral-7B-v0.1 and Qwen2.5-7B as well. Table~\ref{tab:cross_model_core_summary} summarizes the main replication signals. Across both models, subtype searches rediscover nearly all global heads, observed global--family overlap is far above the uniform-head baseline, and additive decomposition changes only minimally after adding interaction terms. Together, these results support the same qualitative conclusion as in LLaMA: a shared global core persists, while subtype-specific residual structure remains comparatively weak.

\begin{table}[H]
\centering
\small
\setlength{\tabcolsep}{5pt}
\begin{tabular}{lcc}
\toprule
\textbf{Metric} & \textbf{Mistral-7B} & \textbf{Qwen2.5-7B} \\
\midrule
Global heads rediscovered & 79/80 & 79/80 \\
Uniform expected overlap & 6.25 & 8.16 \\
Global--family mean overlap & 40.75 & 46.25 \\
Additive \(R^2\) & 0.523 & 0.553 \\
Full-interaction \(R^2\) & 0.528 & 0.560 \\
Prediction correlation & 0.995 & 0.994 \\
\bottomrule
\end{tabular}
\caption{Compact summary of the main cross-model replication results under the same protocol. Overlap rows compare observed head-set overlap against the within-model uniform-head baseline; the final three rows summarize the additive decomposition analysis.}
\label{tab:cross_model_core_summary}
\end{table}

\begin{table}[H]
\centering
\small
\setlength{\tabcolsep}{6pt}
\resizebox{\columnwidth}{!}{%
\begin{tabular}{lccc}
\toprule
\textbf{Model} & \textbf{Diff.-family} & \textbf{Same-family} & \textbf{Same-subtype} \\
\midrule
LLaMA-3.1-8B & 1.000 & 1.141 & 1.675 \\
Mistral-7B   & 1.000 & 1.239 & 1.823 \\
Qwen2.5-7B  & 1.000 & 1.229 & 1.614 \\
\bottomrule
\end{tabular}%
}
\caption{Cross-model transfer summary. Values are non-target-centered normalized recovery under valid-attack filtering. Row-wise normalization by the different-family value is used to reduce apparent differences induced by model-specific scale.}
\label{tab:cross_model_transfer}
\end{table}

\section{External Clean and Context-Use Probes}
\label{app:lre}

\begin{table}[H]
\centering
\small
\begin{tabular}{lcc}
\toprule
\textbf{Circuit} & \textbf{Top-3} & \textbf{Top-10} \\
\midrule
Base & 82.90 & 92.86 \\
Global shared & 84.07 & 92.86 \\
Family-shared avg. & 82.17 & 92.89 \\
Subtype-specific avg. & 78.50 & 91.72 \\
Subtype-specific: country--capital & 72.25 & 90.87 \\
\bottomrule
\end{tabular}
\caption{
External LRE clean capability. Global-shared ablation preserves clean relation completion, while subtype-specific circuits cause larger collateral degradation on average.
}
\label{tab:clean_capability_family_breakdown}
\end{table}

\paragraph{External LRE clean capability.}
We evaluate relation-completion prompts without distractor context.
The global mixed-relation circuit preserves clean relation-completion accuracy, while the subtype-specific circuits cause larger top-3 degradation.
\begin{table}[H]
\centering
\small
\setlength{\tabcolsep}{4pt}
\resizebox{\columnwidth}{!}{%
\begin{tabular}{lcccccc}
\toprule
\textbf{Relation} &
\multicolumn{3}{c}{\textbf{Top-3}} &
\multicolumn{3}{c}{\textbf{Top-10}} \\
\cmidrule(lr){2-4} \cmidrule(lr){5-7}
& \textbf{Base} & \textbf{Global} & \textbf{Subtype} &
\textbf{Base} & \textbf{Global} & \textbf{Subtype} \\
\midrule
fruit outside color & 40.0 & 56.7 & 13.3 & 86.7 & 83.3 & 86.7 \\
landmark in country & 90.0 & 91.5 & 88.0 & 95.5 & 96.0 & 95.5 \\
landmark on continent & 84.5 & 84.0 & 76.5 & 97.0 & 97.5 & 97.0 \\
product by company & 85.5 & 86.0 & 84.0 & 93.5 & 93.0 & 91.0 \\
star constellation name & 85.6 & 87.8 & 41.7 & 89.2 & 89.9 & 80.6 \\
task done by tool & 68.8 & 71.9 & 68.8 & 90.6 & 90.6 & 90.6 \\
task person type & 80.0 & 80.0 & 80.0 & 85.0 & 85.0 & 85.0 \\
work location & 57.6 & 51.5 & 60.6 & 75.8 & 72.7 & 75.8 \\
\midrule
\textbf{Overall} & \textbf{82.9} & \textbf{84.1} & \textbf{72.2} &
\textbf{92.9} & \textbf{92.9} & \textbf{90.9} \\
\bottomrule
\end{tabular}%
}
\caption{External LRE clean capability test (full per-relation breakdown; corresponds to Figure~\ref{fig:clean_capability} in the main text). Base denotes no ablation; Global denotes the mixed-relation global circuit; Subtype denotes the country--capital subtype-specific circuit.}
\label{tab:lre_clean_capability}
\end{table}

\paragraph{Context-use preservation.}
We also evaluate simple context-use tasks, including few-shot spelling
correction, few-shot translation, and synthetic in-context mapping. Global
ablation leaves the spelling and translation probes unchanged at
100\% top-10 accuracy across all tested shot counts. On the harder synthetic
mapping probes, performance is more variable, with the largest degradation on
nonce-label mapping. Overall, top-1 accuracy is unchanged, top-3 accuracy drops
from \(39.71\) to \(38.60\), and top-10 accuracy drops from \(66.91\) to
\(61.03\) (Table~\ref{tab:context_use_preservation}).
\begin{table}[H]
\centering
\scriptsize
\setlength{\tabcolsep}{3pt}
\begin{tabular}{lrrrrrrr}
\toprule
\textbf{Circuit} & \textbf{$n$} & \textbf{$K$} & \textbf{Top-1} & \textbf{Top-3} & \textbf{Top-10} & \textbf{$\Delta$Top-3} & \textbf{$\Delta$Top-10} \\
\midrule
Base & 272 & 0  & 31.62 & 39.71 & 66.91 & 0.00 & 0.00 \\
Global shared & 272 & 80 & 31.62 & 38.60 & 61.03 & 1.10 & 5.88 \\
\bottomrule
\end{tabular}
\caption{Expanded context-use preservation probe. Accuracies and drops are in percentage points; positive drops mean worse than the base model.}
\label{tab:context_use_preservation}
\end{table}

\FloatBarrier
\begin{figure*}[t]
    \centering
    \includegraphics[width=0.95\textwidth]{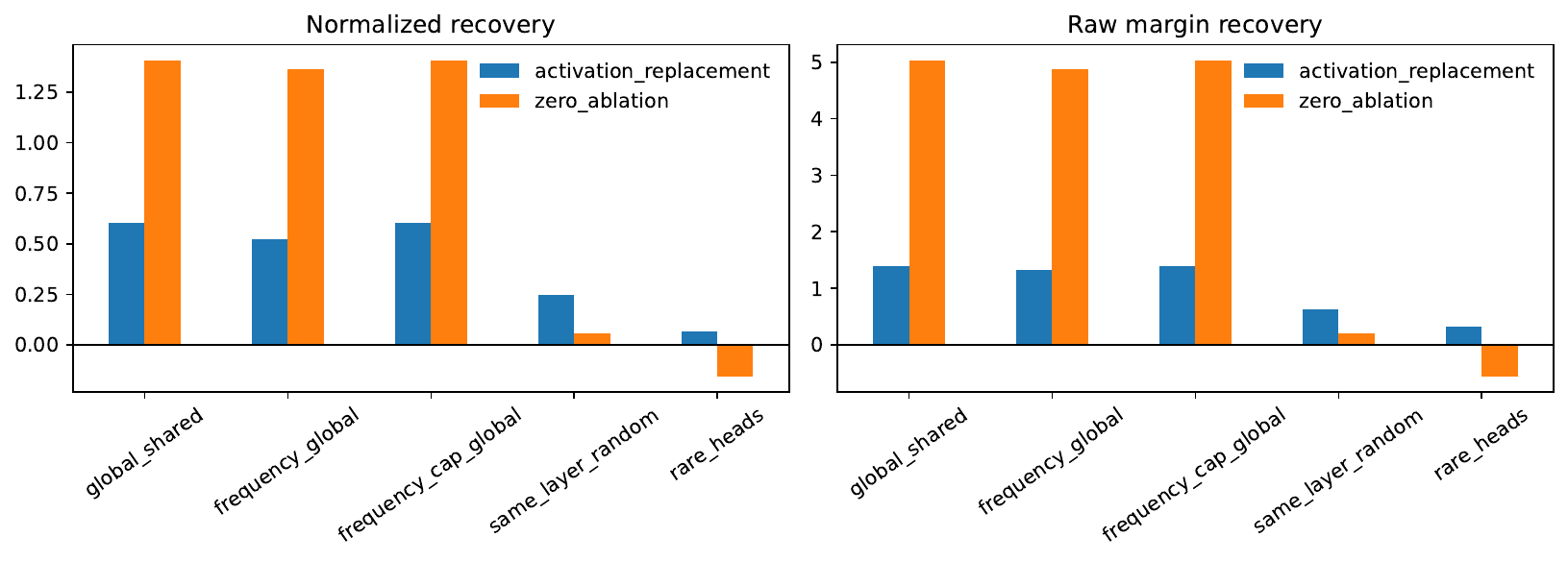}
    \caption{Activation replacement robustness. Clean-to-attacked activation replacement recovers substantial attacked margin for global and frequency-defined global heads, while rare heads remain weak. Zero ablation is stronger, but the replacement result shows that the global component is not merely a zero-ablation artifact.}
    \label{fig:activation_replacement}
\end{figure*}

\section{Activation Replacement Robustness}
\label{app:activation_replacement}

To test whether the global component is specific to zero ablation, we run a clean-to-attacked activation replacement intervention. For each valid attacked prompt, we first run the corresponding clean prompt and cache the selected heads' final-position outputs. We then run the attacked prompt while replacing those same head outputs at the answer position with their clean-prompt activations. This intervention is less destructive than zero ablation and tests whether the selected heads carry attack-sensitive contextual signal.

Figure~\ref{fig:activation_replacement} reports the results on 100 valid attacked cases, balanced across the four semantic families. Global-shared heads recover 60.4\% of the attack-induced margin loss under activation replacement, with a raw margin increase of 1.40. Frequency-defined global heads show a similar effect, recovering 52.4\% of the margin loss. By contrast, rare heads recover only 6.5\%. The logit decomposition further distinguishes global heads from controls: global replacement both increases the gold logit and decreases the distractor logit, whereas rare and same-layer random heads reduce the gold logit. This supports the interpretation that the global component is not merely a zero-ablation artifact.

\FloatBarrier
\section{Additive Decomposition Details}
\label{app:additive_decomposition}

We test whether the mixed-circuit decomposition is approximately additive at the level of causal intervention effects. For each target subtype, we construct disjoint global, family-residual, and subtype-residual components and evaluate interventions that include or exclude each component. We then fit regressions predicting raw margin recovery $\Delta$ from component indicators.

\paragraph{Factorial regression.}
The additive model is
\[
\Delta
=
\beta_0
+
\beta_G I_G
+
\beta_F I_F
+
\beta_S I_S
+
\epsilon,
\]
where \(I_G\), \(I_F\), and \(I_S\) indicate whether the global, family-residual, or subtype-residual component is ablated. Table~\ref{tab:additive_regression} compares this additive model with a full interaction model.

\begin{table}[!ht]
\centering
\small
\setlength{\tabcolsep}{5pt}
\begin{tabular}{lcc}
\toprule
\textbf{Term} & \textbf{Additive} & \textbf{Full interaction} \\
\midrule
Intercept & $-0.254$ & $-0.001$ \\
$G$ & $2.936$ & $2.492$ \\
$F$ & $1.009$ & $0.735$ \\
$S$ & $1.159$ & $0.886$ \\
$G{:}F$ & -- & $0.425$ \\
$G{:}S$ & -- & $0.422$ \\
$F{:}S$ & -- & $0.085$ \\
$G{:}F{:}S$ & -- & $0.079$ \\
\midrule
$R^2$ & $0.572$ & $0.578$ \\
\bottomrule
\end{tabular}
\vspace{-1.5mm}
\caption{Factorial regression on raw margin recovery; interaction terms add little beyond the additive model.}
\label{tab:additive_regression}
\end{table}

The additive model captures most of the explainable structure: adding all interaction terms increases \(R^2\) from \(0.572\) to \(0.578\), and the predictions of the additive and full-interaction models are nearly identical (\(r=0.995\)). Thus, the decomposition should not be read as a literal linear factorization of model internals, but it is a good functional approximation of how the causal effects combine.

\paragraph{Fixed-budget diagnostic.}
As a complementary check, we construct an 80-head mixed intervention by combining \(G_{40}\), \(F_{20}\), and \(S_{20}\): the top 40 global heads, 20 family-residual heads excluding those global heads, and 20 subtype-residual heads excluding both previous sets. This keeps the total intervention size fixed at 80 heads.

\begin{table}[!b]
\centering
\scriptsize
\setlength{\tabcolsep}{4pt}
\resizebox{\columnwidth}{!}{%
\begin{tabular}{lccc}
\toprule
\textbf{Condition} & \textbf{Recovery} & \textbf{Raw $\Delta$} & \textbf{Clean harm} \\
\midrule
$G_{40}+F_{20}+S_{20}$ & $1.550$ & $5.003$ & $-1.012$ \\
$G_{80}$ global only & $1.464$ & $4.900$ & $-0.990$ \\
$S_{80}$ subtype only & $1.404$ & $4.715$ & $-1.073$ \\
$F_{80}$ family only & $1.306$ & $4.254$ & $-1.111$ \\
Random 80 & $-0.091$ & $-0.308$ & $0.523$ \\
Same-layer random 80 & $-0.321$ & $-1.000$ & $0.664$ \\
\bottomrule
\end{tabular}%
}
\vspace{-1.5mm}
\caption{Fixed-budget diagnostic (80 heads): the structured mix slightly outperforms single-component circuits; random controls fail.}
\label{tab:fixed_budget_gfs}
\end{table}

This fixed-budget result is diagnostic rather than an optimized allocation. Its purpose is to show that family- and subtype-residual components can add useful signal beyond the global core under a controlled head budget. Combined with the regression analysis, it supports the mixed-circuit interpretation at the level of causal intervention effects: \(G\) is the dominant component, while \(F_{\mathrm{fam}(r)}\) and \(S_r\) provide smaller positive residual effects.

\FloatBarrier
\section{Illustrative Gumbel-Sigmoid Intervention}
\label{app:gumbel_example}

\Needspace{10\baselineskip}
\begin{figure}[!t]
    \centering
    \includegraphics[width=\columnwidth,height=0.45\textheight,keepaspectratio]{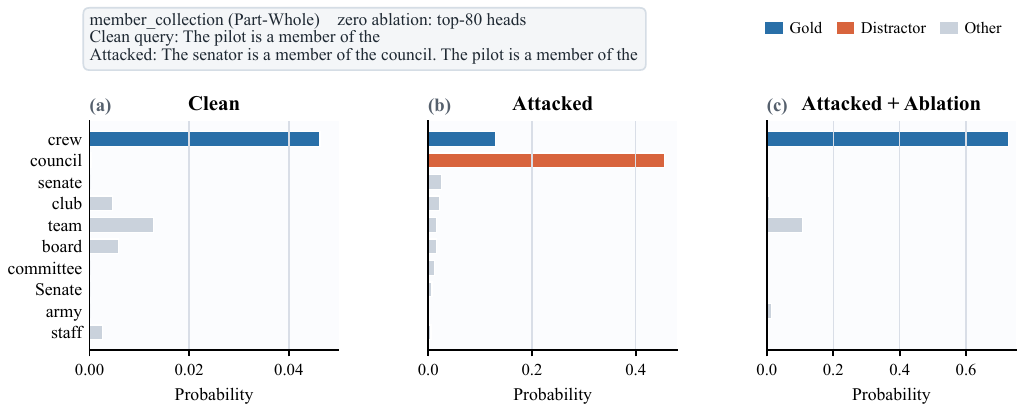}
    \caption{An illustrative factual-distractor example under the ranking-based intervention used in our Gumbel-Sigmoid head search. The clean prompt favors \texttt{crew}; the distractor context shifts probability toward \texttt{council}; and zero-ablating the top-80 heads exported from the learned mask scores reverses this shift.}
    \label{fig:gumbel}
\end{figure}

\noindent Figure~\ref{fig:gumbel} shows a representative valid attacked case used to visualize the intervention produced by our Gumbel-Sigmoid head search.
The figure is illustrative rather than a separate evaluation result; all quantitative analyses in the main text are computed over filtered held-out attack cases.
\FloatBarrier
\section{Gumbel-Sigmoid Search Details}
\label{app:gumbel_details}

This appendix provides additional details for the differentiable head-search procedure described in Section 3.2. For each attention head \(h_i\), we introduce a learnable mask logit \(\alpha_i\). During search, we sample a relaxed mask
\[
\tilde{m}_i
=
\sigma\left(
\frac{\alpha_i + g_i}{\tau}
\right),
\]
where \(g_i\) is logistic noise induced by the Gumbel reparameterization and \(\tau\) is a temperature hyperparameter. In the forward pass, we use a straight-through binary mask obtained by thresholding \(\tilde{m}_i\), while gradients are passed through the relaxed value. A binary value of \(m_i=1\) means that the corresponding head output is zero-ablated; \(m_i=0\) leaves the head active.

For a batch of attacked prompts \(\mathcal{B}\), the search objective is
\begingroup
\small
\[
\mathcal{L}
=
-
\mathbb{E}_{x\in\mathcal{B}}
\left[
M_{\theta}^{\mathrm{abl}}(x)-M_{\theta}(x)
\right]
+
\frac{\lambda}{|\mathcal{H}|}
\sum_{i=1}^{|\mathcal{H}|}
\sigma(\alpha_i),
\]
\endgroup
where \(M_{\theta}(x)\) is the attacked margin of the original model, \(M_{\theta}^{\mathrm{abl}}(x)\) is the margin under the current differentiable mask, \(\mathcal{H}\) is the set of all attention heads, and \(\lambda\) controls sparsity. Since \(M_{\theta}(x)\) is constant with respect to the mask for a fixed batch, this objective is equivalent during optimization to maximizing the ablated margin while penalizing the expected number of selected heads.

We anneal the Gumbel-Sigmoid temperature during search and optimize mask logits with gradient descent while keeping all model parameters frozen. After optimization, the learned mask scores define a ranking over heads, and the main experiments use the top \(K=80\) interventions exported from that ranking. We select this value with a held-out \(K\)-sweep over \(K\in\{20,40,60,80\}\), choosing a setting that provides strong attacked-margin recovery while avoiding excessive clean-prompt harm. For frequency-defined analyses, we additionally report results at \(K\in\{20,40,80\}\).

For the main LLaMA-3.1-8B experiments, subtype-specific Gumbel searches use 450 optimization steps, while family-shared and global-shared searches use 650 steps because they aggregate more heterogeneous training cases. The seed-stability runs use 300 steps. We use a search batch size of 2, an evaluation batch size of 4, learning rate \(0.08\), sparsity weight \(\lambda=0.025\), and a Gumbel-Sigmoid temperature annealed from \(1.0\) to \(0.25\). The main head budget is \(K=80\), selected from a held-out sweep over \(K\in\{20,40,60,80\}\). Each subtype uses 120 train attack pairs and 120 test attack pairs before valid-attack filtering; the main valid-attack threshold is \(D\geq0.5\).

\section{Artifact use.}
We use publicly available model checkpoints and relation-completion resources for research evaluation, following their respective access conditions. Our constructed benchmark consists of synthetic templates and factual lexical-semantic pairs.
\end{document}